\documentclass[sigconf,nonacm]{acmart}

\AtBeginDocument{%
  }

\begin{document}

\title{Using Weak Supervision and Data Augmentation in Question Answering}

\author{Chumki Basu}
\affiliation{%
  \institution{Peraton Labs}
  \streetaddress{150 Mount Airy Road}
  \city{Basking Ridge}
  \state{New Jersey}
  \country{USA}
  \postcode{07920}
}
\email{cbasu@peratonlabs.com}

\author{Himanshu Garg}
\affiliation{%
  \institution{Peraton Labs}
  \streetaddress{150 Mount Airy Road}
  \city{Basking Ridge}
  \state{New Jersey}
  \country{USA}
  \postcode{07920}
}
\email{himanshu.garg@peratonlabs.com}

\author{Allen McIntosh}
\affiliation{%
  \institution{Peraton Labs}
  \streetaddress{150 Mount Airy Road}
  \city{Basking Ridge}
  \state{New Jersey}
  \country{USA}
  \postcode{07920}
}
\email{aamcintosh@peratonlabs.com}

\author{Sezai Sablak}
\affiliation{%
  \institution{Peraton Labs}
  \streetaddress{150 Mount Airy Road}
  \city{Basking Ridge}
  \state{New Jersey}
  \country{USA}
  \postcode{07920}
}
\email{sezai.sablak@peratonlabs.com}

\author{John R. Wullert II}
\affiliation{%
  \institution{Peraton Labs}
  \streetaddress{150 Mount Airy Road}
  \city{Basking Ridge}
  \state{New Jersey}
  \country{USA}
  \postcode{07920}
}
\email{jwullert@peratonlabs.com}

\renewcommand{\shortauthors}{Basu et al.}

\begin{abstract}
The onset of the COVID-19 pandemic accentuated the need for access to biomedical literature to answer timely and disease-specific questions. During the early days of the pandemic, one of the biggest challenges we faced was the lack of peer-reviewed biomedical articles on COVID-19 that could be used to train machine learning models for question answering (QA). In this paper, we explore the roles weak supervision and data augmentation play in training deep neural network QA models. First, we investigate whether labels generated automatically from the structured abstracts of scholarly papers using an information retrieval algorithm, BM25, provide a weak supervision signal to train an extractive QA model. We also curate new QA pairs using information retrieval techniques, guided by the \emph{clinicaltrials.gov} schema and the structured abstracts of articles, in the absence of annotated data from biomedical domain experts.  Furthermore, we explore augmenting the training data of a deep neural network model with linguistic features from external sources such as lexical databases to account for variations in word morphology and meaning. To better utilize our training data, we apply curriculum learning to domain adaptation, fine-tuning our QA model in stages based on characteristics of the QA pairs. We evaluate our methods in the context of QA models at the core of a system to answer questions about COVID-19.
\end{abstract}



\maketitle

\section{Introduction}

\noindent 
Machine learning models typically benefit from a wealth of training data to achieve state-of-the-art performance on tasks such as question answering \cite{rzll:16}. However, during the early stages of the COVID-19 pandemic, we were dealing with a paucity of topical scholarly articles that were available for model training, and an even smaller set that was labelled by domain experts for the question-answering task. This shaped our research perspective, which is driven primarily by how to implement cost-effective methods for generating training data. Towards this objective, we focus on weak supervision and data augmentation. We also explore how to better utilize this training data by applying curriculum learning to domain adaptation for question answering.

Machine learning practitioners often explore the space of model architectures to find the best one that fits the data -- our objective is to fix a baseline model architecture and focus on varying the training data. The lack of attention paid to the training process has been pointed out in the recent literature \cite{sirs:22}, and consequently, we are addressing an important need. Specifically, in our variations of the training data, we investigate the role structured information could play in labelling repurposed data, curating data, and in augmenting data with linguistic features. While information retrieval has an established role in QA systems, we also explore the role information retrieval could play in labelling and curating data, particularly with the help of structured data.

PubMed \cite{pu:22} is a resource that provides access to an extensive collection of scholarly biomedical articles (36 million citations as of this writing). In the early part of our work, we found a publicly available biomedical QA dataset called PubMedQA \cite{p:22,jdlcl:19}, compiled from PubMed abstracts. PubMedQA \cite{jdlcl:19} consists of QA pairs derived from the automatic collection and manual annotation of scholarly PubMed articles biased towards clinical study-related topics. That work \cite{jdlcl:19} also established a baseline for yes/no/maybe QA using the BioBERT \cite{lykkksk:20} neural language model. However, \citeauthor{jdlcl:19} (\citeyear{jdlcl:19}) had not tackled extractive QA.

We repurpose PubMedQA \cite{jdlcl:19} for extractive QA. Relying on structured PubMed abstracts from this dataset, we extract text from a part of an article's abstract using the BM25 \cite{rz:09} algorithm and return the extracted text as an answer to a query. While BM25-based weak supervision enables us to generate the training data needed to stand-up a system that could field biomedical queries, we also augment the dataset with COVID-specific articles. We tackle the challenge of data curation by formulating it as two information retrieval tasks guided by knowledge -- in one instantiation, we use knowledge in the form of external schema (e.g., from \emph{clinicaltrials.gov} \cite{x:21}) in query formulation, and in a second instantiation, we use the structure of article abstracts in results selection. Collectively, our data can be decomposed into subsets with different characteristics. From the perspective of model training, we stage the fine-tuning process from broad, biomedical questions to COVID-specific questions, and for the COVID-specific questions, we stage the fine-tuning process from easy to hard questions.

We also investigate transforming the training data of our models using linguistic features from external sources such as lexical databases to smooth out differences in word morphology and meaning. We investigate data augmentation outside of the extractive QA system, in the context of yes/no QA, enabling us to compare our results against the PubMedQA experiments \cite{jdlcl:19}.

The contributions of this paper are:

\begin{enumerate}
\item Training an extractive QA model using BM25-based weak supervision with labels generated automatically from repurposed structured article abstracts.
\item Curating COVID-specific QA pairs comprised of natural and ``artificial'' questions using information retrieval techniques guided by structured  knowledge, in the absence of annotations from domain experts.
\item Applying curriculum learning to specialize a domain for extractive QA.
\item Developing multiple techniques to augment the training data of a QA model with linguistic features from external sources such as lexical databases to account for variations in word morphology and meaning and combining these techniques to improve model performance on yes/no QA.
\end{enumerate}

\section{Background}
\noindent In our work, we explore two question answering (QA) tasks:

\begin{itemize}
\item Extractive QA: The answer in a QA pair is a sentence extracted from some context.
\item Yes/no QA: The answer in a QA pair is a ``yes'' or ``no'' label.
\end{itemize}

In extractive QA, we learn a function that takes a question as input and outputs a sentence-level answer from some context, e.g., the ``conclusions'' section of a paper abstract. In yes/no QA, we learn a function that takes a question as input and outputs one of two labels, ``yes'' or ``no''. (Note that this is a simplification of the yes/no/maybe QA problem discussed in the PubMedQA paper \cite{jdlcl:19}). There has been much prior work tackling challenges related to retrieving information and answering questions in the biomedical domain. For background, we point the reader to survey papers \cite{ah:10,jyxyytchly:22} as well as the approaches of an early clinical QA system \cite{clsabcey:11}.

PubMedQA \cite{jdlcl:19} consists of labelled data (PQA-L), unlabelled data (PQA-U), and data whose questions and labels were generated artificially (PQA-A). PQA-U consists of PubMed articles having a question mark in the titles and a structured abstract with a ``conclusions'' section. Two annotators then hand-labelled 1000 instances from PQA-U with yes/no/maybe labels to form PQA-L. PQA-A consists of 211.3K ``noisily labelled instances'' with yes/no labels where titles (1) satisfy a pre-defined part-of-speech-tagging structure and (2)  abstracts contain a ``conclusions'' section. The ``conclusions'' section forms the basis of a ``long answer'' to a question; for extractive QA, we map this label directly to the equivalent label marking, ``context''.
 
We choose BioBERT (Bidirectional Encoder Representations from Transformers for Biomedical Text Mining) \cite{lykkksk:19,lykkksk:20}, a domain-specific language representation model pre-trained on large-scale biomedical corpora (PubMed abstracts and PubMed Central full-text papers), as our baseline model architecture \cite{bio:20}. BioBERT builds on BERT \cite{dclt:18}, a bidirectional, language representation model that incorporates context from both directions (left and right). BERT was trained and tested primarily on datasets comprised of general domain texts. The BioBERT model is initialized with pre-trained parameters, and during model fine-tuning, we fine-tune these parameters using labelled data from a question-answering task. While there are other models pre-trained on PubMed texts that have since shown improvement by domain-specific pre-training from scratch such as PubMedBERT \cite{gtclulngp:21}, when we started our work, BioBERT was a leading model and was used in the PubMedQA \cite{jdlcl:19} experiments, which facilitates direct comparison. We note that the baseline model architecture was pre-trained on SQuAD v1.1 \cite{rzll:16} on top of BioBERT v1.1 \cite{bio:20,lykkksk:19}.

We evaluate our extractive QA models using:
\begin{itemize}
\item Exact match score: The percentage of predicted answers that match any of the ground truth answers exactly.
\item $F1$-score: The average word overlap between the predicted answer and the ground truth answer.
\end{itemize}
For yes/no QA, we evaluate our models using accuracy -- the percentage of correct predictions. \citeauthor{jdlcl:19}  (\citeyear{jdlcl:19}) randomly sampled 500 PQA-L instances for 10-fold cross-validation and set aside the remaining 500 instances as the test set. Consequently, we also perform 10-fold cross validation with 450 training instances in each fold.

\section{Methodology}
In Figure 1, we show our QA system architecture. The QA Management Function accepts a question from the User Interface, submits it to the Retrieval Function, implemented using $ElasticSearch$ \cite{g:15}, and receives potentially relevant documents in response.      The $ElasticSearch$ index is populated with a set of PubMed abstracts that contained at least one of the phrases ``COVID'', ``SARS-CoV-2'' or ``novel coronavirus''. The QA Management Function then submits question/document pairs to the Answer Generation Function and receives answers in response. The Answer Generation Function queries the best-performing extractive QA model as determined by the fine-tuning scenarios.

This modular design aids extensibility -- e.g., currently, we support multiple extractive QA models, allowing the user to toggle between them in the User Interface. In the future, we plan to integrate yes/no QA models. 

\begin{figure}[h]
  \centering
  \includegraphics[width=\linewidth]{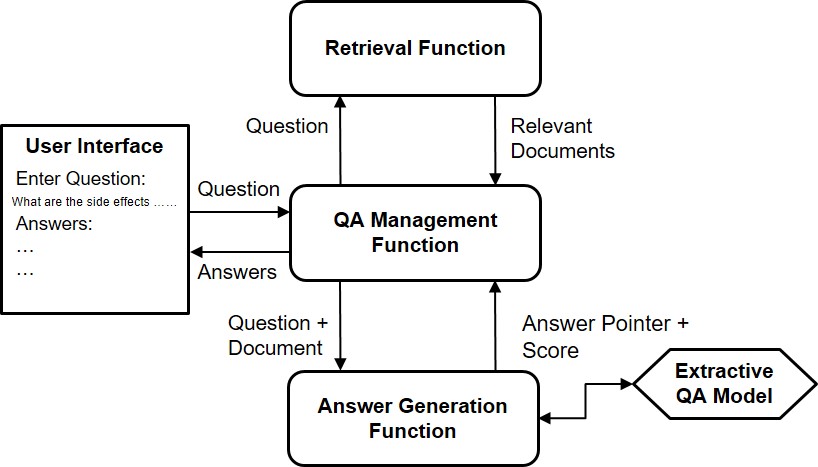}
  \caption{QA system architecture.}
  \Description{QA system architecture.}
\end{figure}

In the following sections, we describe our techniques for varying the training data, first, in the context of the ``Extractive QA Model''. Then, we show how to vary training data in the context of a ``Yes/No model'' that is yet to be integrated into our system architecture.

\subsection{Generating labelled QA pairs for weak supervision}
For the extractive QA task, we provide our learning model some context (i.e., the ``conclusions'' in an article's abstract) and the model learns to extract an answer (i.e., sentence) from that context. To facilitate understanding of the results of extractive QA, we return the answer sentence highlighted in the associated context as the final result of QA.

\citeauthor{jdlcl:19}  (\citeyear{jdlcl:19}) created data subsets, PQA-L and PQA-A, assigning yes/no/maybe and yes/no ground truth labels, respectively, as described earlier. For our work, the existence of these subsets suggested that, for a cross-section of PubMed papers, the articles' abstracts contained sufficient information to answer the questions posed by their titles. Our job is to find an algorithm to extract answers, automatically, from PQA-L and PQA-A. We choose BM25 \cite{rz:09}, an information retrieval algorithm widely used by search engines to rank the similarity of documents to queries, and assign ground truth labels as follows:

\begin{enumerate}
\item Label an article's ``conclusions'' as ``context''.
\item Use BM25 to find the sentence in an article's ``conclusions'' that best matches the ``question'' and label that sentence as the ``answer''.
\end{enumerate}

We refer to the repurposed data subsets with new ground truth and ``context'' labels as PQA-L$'$ and PQA-A$'$.  While BM25 is a standard information retrieval algorithm, one could expect to find it in the ``Retrieval Function'' of a QA system (Figure 1). In early Open-Domain QA systems, BM25 was used to retrieve candidate passages relevant to a user's question, though it was noted that BM25's notions of relevance are not tailored to questions \cite{kpz:21}. However, if one were to consider a ranking task instead of QA, BM25 labelling has been used, effectively, as a weak supervision signal to train a neural ranker \cite{dzskc:17}. In that work, \citeauthor{dzskc:17} (\citeyear{dzskc:17}) started with a large set of training queries and a target collection of documents without relevance judgments and used a BM25 ``pseudo-labeller'', which provided a weak supervision signal, to score the relevance of each document with respect to each query. The resulting labelled data set was then used to train a ranking model. 

Our novelty is using BM25 labelling as a weak supervision signal to train an extractive QA model. We also acknowledge that BM25 is an imperfect algorithm for the answer extraction task -- BM25 will identify a sentence that is highly similar to a question, but that sentence may not necessarily contain an answer to the question. If an abstract does contain an answer to the original question, we assume that it is more likely to appear in the ``conclusions'' section of an abstract rather than the other sections. Therefore, the knowledge captured in an article's structured abstract justified our directed use of BM25 to extract an answer.

To evaluate BM25-labelling performance, we manually labelled a small, random sample of the data; inter-annotator agreement between the human labels and BM25 sentence-level answers measured using the exact match score was 80\%. We evaluate the efficacy of BM25 at extractive QA by fine-tuning our model on different training schedules.

\subsection{Curating COVID-specific QA pairs}
Since PubMedQA \cite{jdlcl:19} was created before the pandemic, we face the task of curating QA pairs relevant to COVID-19. Curation consists of two primary steps: retrieving relevant, candidate QA pairs and extracting answers from the candidates. There may be an additional filtering step that prunes candidate pairs that did not contain the answer to the question. We consider representing both natural questions as well as the type of artificial questions that comprised PQA-A in our curated dataset. 
 
By natural questions, we mean questions that we anticipate will be asked by end users in search queries, on FAQs, etc. However, as a precursor to tackling the more challenging problem of free-form natural language queries, we start by constructing parameterized question templates, where instances vary based on one or more typed parameters. To constrain the values of the parameters, we use terms from the \emph{clinicaltrials.gov} schema as well other controlled vocabularies. Since our objective is to extract answers from retrieved articles that present clinical trial results, we use the \emph{clinicaltrials.gov} schema to map the language of queries to the language of papers that discussed clinical trial results.

We identify question templates of the form:
\begin{itemize}
\item What risks does a person with [CONDITION] face with respect to COVID-19?
\item What adverse effects are associated with [TREATMENT]?
\end{itemize}
Our expectation is that by fine-tuning our model with multiple QA instances for each question matching a given template, the model would learn to answer questions of a similar form with other parameter values.

We show examples of values assigned to parameters below:
\begin{itemize}
\item Risk Conditions: asthma, cardiovascular disease, diabetes, kidney disease, obesity
\item Adverse Effect Treatments: Azithromycin, Dexamethasone, Hydroxychloroquine, Infliximab, Ivermectin, Tocilizuma
\item Adverse Effect Treatment/Condition Pairs: Dexamethasone/ arthritis, Dexamethasone/post-operative nausea, Dexamethasone/ chemotherapy-induced nausea
\end{itemize}

We generate the dataset using the following process:
\begin{enumerate}
\item Choose question template and populate an instance with a set of parameters.
\item Use parameters from a question instance to formulate a PubMed search query and retrieve PubMed abstracts addressing parameters.
\item Select abstracts where answer appears in Results/Conclusion sections.
\end{enumerate}

The generation of parameterized QA pairs includes a manual answer identification (selection) step to ensure high data quality. Unlike other annotated datasets \cite{mrjp:20,tbmpzawkppapbganhgbsap:16,b:21}, the manual review is not performed by a biomedical domain expert. The novelty of our template-based approach is that the specification of parameters provides knowledge in the form of a focused set of keywords that not only guides structured query formulation, but also, aids the non-domain-expert reviewer in locating answers to questions in the retrieved abstracts.

For each instance, we identify answers from multiple published abstracts (50), capturing both answer sentence and associated context. The resulting data set is not large (more than 500 QA were used to fine-tune our model) but contains multiple answers to each question. Prior to model fine-tuning, we held out one adverse-effect-treatment set to evaluate generalization.

For an example parameterized question, \emph{What risks do patients with diabetes face with respect to COVID-19?}, we manually identify an answer sentence (italicized) in its associated context:
\begin{quote}
\emph{There is evidence of increased incidence and severity of COVID-19 in patients with diabetes.} COVID-19 could have effect on the pathophysiology of diabetes. Blood glucose control is important not only for patients who are infected with COVID-19, but also for those without the disease. Innovations like telemedicine are useful to treat patients with diabetes in today's times.
\end{quote}

Curation of templates builds on prior work in the biomedical literature on ``what'' and ``how''-type questions \cite{clsabcey:11}. However, we do understand the limitations of template-based approaches and acknowledge that we will need to increase the diversity of parameterized COVID QA pairs to handle variation in clinical questions. 

We also curate QA pairs from PubMed papers, automatically, by finding matched pairs of schema labels for questions and answers in PubMed abstract sentences. Like PQA-A, these QA pairs are ``noisily labelled instances'', not manually labelled or edited in any way. However, given the limited nature of data in a pandemic, this data is orders of magnitude smaller -- approximately, 1K curated instances compared to 211.3K instances in PQA-A. 

To generate the dataset, first, we create the ``COVID'' set of PubMed abstracts. Of the unique PubMed papers downloaded, we identify COVID-specific papers by extracting all abstracts containing the phrases ``COVID'', ``SARS-CoV-2'' or ``novel coronavirus'' (case insensitive) in the abstract text, the abstract labels, or the keywords. We ignore papers published before November 1, 2019. This dropped most papers that referred to ``SARS-CoV-1'' as the ``novel coronavirus''. 

If the abstract has a labelled ``Question'' section, we use that as the ``question'' of a QA pair. Otherwise, if the paper title contains a question mark, we used that as the ``question''. We then check to see if the abstract had a ``Conclusion'' or ``Answer'' section -- i.e., we take any sentence with a label containing an ``ANSWER'' or ``CONCLUSION(S)'' (including a misspelling of the latter), provided it does not also contain, ``RESULT'', ``SUMMARY'', ``FINDING'', or ``DISCUSSION''. Finally, we apply the BM25-labelling steps as described earlier to extract a sentence-level answer for the QA pair. We note that there is no manual review of the labelling results.

\subsection{Learning from easy to hard examples}
We expect QA model performance to degrade when the target distribution differs from the source distribution. To address the challenge of domain adaptation, in a supervised setting, prior work \cite{yzksw:22,kjl:20,snznnwnx:20,ykf:21} has shown that integrating labeled target QA pairs during training improves performance on out-of-domain questions. This is true when the target data is either human-annotated QA pairs or synthetic data derived from question generation techniques. Furthermore, \citeauthor{zskmcd:19} (\citeyear{zskmcd:19}) have shown that the training process can be divided into phases, where earlier training phases incorporate easier examples. 

\citeauthor{zskmcd:19} (\citeyear{zskmcd:19}) applied curriculum learning to domain adaptation and demonstrated its effectiveness on neural machine translation models. In related literature, \citeauthor{blcw:09} (\citeyear{blcw:09}) and \citeauthor{sirs:22} (\citeyear{sirs:22}) present curriculum learning as a method that gradually increases the complexity of training examples, relying on an easy-to-hard ranking of the examples. There are many flavors of curriculum learning, which have been applied across a number of tasks in machine learning. These instantiations of curriculum learning include the employment of staged scheduling, where each stage focused on a specific task \cite{lsc:17,nsls:16,zdg:17,sirs:22}. \citeauthor{ctlmce:19} (\citeyear{ctlmce:19}) employed a multi-stage learning approach with data ordered from the most semantically generic to most specific for the spoken language understanding task. Our work is similar to curriculum learning along two dimensions: we fine-tune a model, progressively, from (a) broad to specific questions, and for the specific questions, we fine-tune a model from (b) easy to hard questions. 

Following \citeauthor{jdlcl:19} (\citeyear{jdlcl:19}), we devise training schedules for multi-phase model fine-tuning in stages. \citeauthor{jdlcl:19} (\citeyear{jdlcl:19}) demonstrated that fine-tuning a model, BioBERT, sequentially, on different subsets of data outperformed single-phase fine-tuning due to different properties of the data subsets. \citeauthor{xeywvm:18} (\citeyear{xeywvm:18}) also illustrated the advantages of progressive, multi-phase fine-tuning for domain adaptation, where a model pre-trained on out-of-domain data is gradually fine-tuned on in-domain data. According to \citeauthor{xeywvm:18} (\citeyear{xeywvm:18}), at each stage of sequential fine-tuning, the similarity between the current and target domain is increased, thereby enabling the model to potentially achieve a better fit with the target distribution. We show that staged fine-tuning is also advantageous as we gradually \emph{specialize} a domain according to characteristics of the data.

We fine-tune our BioBERT model, sequentially, on data subsets, $d_i$: $d_1$ = PQA-L$'$, $d_2$ = PQA-A$'$, $d_3$ = Parameterized COVID QA pairs, $d_4$ = PubMed COVID QA pairs. Neither $d_1$ nor $d_2$ has any COVID-specific questions, but both have very diverse questions of a scholarly, scientific nature. Since scientific publications tend to address under-explored and unexplored issues, we do not see a lot of repetition in paper titles. We also see variability in how the titles (and hence, questions) are constructed. On the other hand, natural questions are often repetitive and share other regularities. For example, parameterized questions are similar in structure. The union of these data subsets is $D$. While we fine-tune and evaluate our models on a mix of question types, $D$, is highly skewed towards scholarly questions derived from paper titles.

For evaluation, we held out instances from the following subsets to create a target test set, $T$:
\begin{displaymath}
	 T = (10\% of d_1) \cup (10\% of d_3) \cup (10\% of d_4)
\end{displaymath}
Our target distribution is representative of the kinds of real-world data that we expect to see. We expect there to be natural questions about COVID-19; these questions are represented by our parameterized dataset. We also expect some scholarly questions that probe the scientific literature; these types of questions are more diverse and are reflected in the COVID-specific QA pairs automatically collected from PubMed papers, and more generally, PQA-L$'$. The rest of the data in $D$ is used for model training. 

We created the following sequential fine-tuning schedules for $d_i$:

\begin{itemize}
\item Schedule 1: Fine-tune model on $d_1$ only
\item Schedule 2: Fine-tune model on $d_2$, then on $d_1$
\item Schedule 3: Fine-tune model on $d_2$, then on $d_1$, then on $d_3$
\item Schedule 4: Fine-tune model on $d_2$, then on $d_1$, then on $d_3$, then on $d_4$
\end{itemize}

 From the perspective of domain specialization, we fine-tune our model on broad biomedical questions ($d_2$ and $d_1$), first, and then, we fine-tune on COVID-specific questions ($d_3$ and $d_4$). Furthermore, for COVID-specific questions, we fine-tune our model according to the difficulty of the question type. Specifically, we fine-tune our model on easier, parameterized, natural questions ($d_3$), first, and then, we fine-tune on harder, artificial questions ($d_4$). We also found multi-phase fine-tuning of a model to be a useful tool in measuring the impact of varying the training data subsets on model performance after each stage.

\subsection{Augmenting training data with linguistic features}
In our work, we vary the data used to train our models based on characteristics of the data. Given a training dataset, we also consider alternate representations of the original text, accounting for variations in word morphology and word meaning. Our work is similar to related work in text data augmentation \cite{bkr:22} that standardize data with the goal of increasing the model's generalization capabilities. For pre-trained language models (such as BERT), \citeauthor{lwd:20} (\citeyear{lwd:20}) hypothesized that pre-training provided benefits similar to that achieved by augmentation. \citeauthor{lwd:20} (\citeyear{lwd:20}) conjectured that augmentation techniques would help if they introduced new linguistic patterns unseen during model pre-training and suggested text editing as a techique to vary linguistic structure. Since structured knowledge often exists in external sources such as databases that are not tapped during pre-training, we were motivated to incorporate linguistic patterns from such sources.

We designed novel text editing techniques that vary linguistic structure by injecting morphological forms -- i.e., lemmas and neo-classical combining forms -- into the original text. Lemmas are base (or dictionary) forms of words, minus inflectional endings. Neo-classical forms are word constituents or parts (prefixes, roots, terminals) derived from Latin and Greek \cite{n:21,mbm:88,lgl:13}. Both lemmas and neo-classical combining forms complement BERT's subword tokenization using the WordPiece algorithm \cite{dclt:18}. 

Related work on adversarial training \cite{tjks:20} investigated the use of word-level perturbations in inflectional morphology to reveal linguistic training biases in models such as BERT. In that work, \citeauthor{tjks:20} (\citeyear{tjks:20}) selected the inflected form that maximized the target model's loss for each noun, verb, or adjective in a sentence to generate an adversarial training set. They showed that fine-tuning the models for one epoch on the adversarial training set was sufficient to achieve model robustness on inflectional variations.

There has been promising related work \cite{wwn:17,hcoh:19} on enriching the representational space of neural network models. We focus on enriching the embedding layer of BERT by injecting linguistic features, $X$, from structured sources. Rather than adopting a single approach, we explore a number of techniques, including in-line text editing (that does not increase the size of the training set) as well as the generation of new, transformed training examples, and $combine$ the best performing methods. We distinguish new example generation from in-line text editing by referring to the former as $Augmentation$ since augmentation traditionally results in new training examples. Collectively, we refer to these perturbations as text transformation techniques:

\begin{itemize}
\item Replacement: Replace original text with $X$
\item Concatenation: Concatenate $X$ to original text
\item Augmentation: Generate a new training example in which original text is replaced with $X$
\end{itemize}

As in \cite{tjks:20}, we would like to preserve semantics as much as possible through our transformations. The techniques, $Replacement$ and $Augmentation$, ensure that we do not change the position of the replacement and original text. For $Concatenation$, we append the new text next to the position of the original text. For each of the above techniques, we subsistute a morphological form -- either lemma or neo-classical form -- for $X$. While Wei and Zou \cite{wz:19} also perform text editing, not all of their methods are semantics-preserving. 

Using a lemmatizer, we return lemmas or the base (or dictionary) forms of words, minus inflectional endings. We show an example of how applying each of the above techniques to word lemmas (italicized) transforms the text of the original sentence: 

\begin{enumerate}
\item Original sentence: The ability to recognize different optotypes differs even if their critical details appear under the same visual angle.
\item Lemmas replaced: The ability to recognize different \emph{optotype differ} even if their critical \emph{detail} appear under the same visual angle.
\item Lemmas concatenated: The ability to recognize different optotypes \emph{optotype} differs \emph{differ} even if their critical details \emph{detail} appear under the same visual angle.
\item Lemmas augmented: (1) Original sentence + (2) Sentence with lemmas replaced
\end{enumerate}

For neo-classical compounds, we introduce morphological variations and meaning substitution -- replacing $X$ with combining forms and the meanings of those forms. The processing of neo-classical compounds is relevant for standardizing disease names \cite{lgl:13}. According to \citeauthor{lgl:13} (\citeyear{lgl:13}), we find disease names mentioned in text with morphological and orthographic variations. In addition, synonyms may be used for disease names. We explore some of these variations in our transformations. For example, we expand neo-classical compounds with their constituent forms (prefixes, roots, and terminals). Then, rather than explore synonym usage, we substitute meanings for these constituents.

From NIH SPECIALIST Lexical Tools, we used Neo-classical Combining Forms 2021 Release \cite{n:21}. The NC.DB database contains morphemes used to form neo-classical compounds. The format of this data is ``MORPHEME|MEANING|TYPE'', where types include prefixes, roots, and terminals. We manually augment this data with entries from Global RPH Medical Terminology.

We expand terms to meanings in two stages. First, we identify word parts matching the pattern:

\begin{quote}
(optional)prefix, one or two roots, (optional)terminal
\end{quote}

\noindent Then, we replace word parts with meanings. In other words, for the term, ``dacryoadenitis'', we perform two mappings to arrive at meanings:
\begin{itemize}

\item dacryoadenitis $\rightarrow$ Root[dacryo] Root[aden] Term[itis] 
\item Root[dacryo] Root[aden] Term[itis] $\rightarrow$ tear gland inflammation
\end{itemize}

For our experiments, we start with BioBERT, initialized with its original Transformer \cite{vspujgkp:17} weights, and replicate the fine-tuning schedule from PubMedQA \cite{jdlcl:19}: fine-tune model, sequentially, on PQA-A, and then, on PQA-L. This is our baseline model for yes/no QA. Each text transformation technique results in a modified representation of the input data subsets, PQA-L and PQA-A, which we then use to fine-tune our models. Following \citeauthor{jdlcl:19}'s (\citeyear{jdlcl:19}) multi-phase fine-tuning approach, we develope several sequential fine-tuning schedules as part of our experimental design:

\begin{itemize}
\item Schedule I: Fine-tune model on PQA-A, then on PQA-L, with lemmas replaced
\item Schedule II: Fine-tune model on PQA-A, then on PQA-L, with lemmas concatenated
\item Schedule III: Fine-tune model on PQA-A, then on PQA-L, with lemmas augmented
\item Schedule IV: Fine-tune model on PQA-A, then on PQA-L, with neo-classical forms replaced
\item Schedule V: Fine-tune model on PQA-A, then on PQA-L, with neo-classical forms concatenated
\item Schedule VI: Fine-tune model on PQA-A, then on PQA-L, with neo-classical forms augmented
\item Schedule VII: Fine-tune model on PQA-A, then on PQA-L, with neo-classical form meanings replaced
\item Schedule VIII: Fine-tune model on PQA-A, then on PQA-L, with neo-classical form meanings concatenated
\item Schedule IX: Fine-tune model on PQA-A, then on PQA-L, with neo-classical form meanings augmented
\end{itemize}

In addition to single transformations, we also explore the compounding effect of applying multiple text transformations. To reduce the number of combinations, we introduce three additional fine-tuning schedules that combine the best performing schedules for lemmas, neo-classical forms, and neo-classical form meanings:

\begin{itemize}
\item Schedule X: Fine-tune model on PQA-A, then on PQA-L, with lemmas concatenated and neo-classical forms replaced
\item Schedule XI: Fine-tune model on PQA-A, then on PQA-L, with neo-classical forms replaced and neo-classical form meanings replaced
\item Schedule XII: Fine-tune model on PQA-A, then on PQA-L, with lemmas concatenated, neo-classical forms replaced, and neo-classical form meanings replaced
\end{itemize}
\noindent If a schedule (combination) does not include a transformation involving neo-classical forms, there is no expansion of the forms into meanings.

\section{Results and discussion}
\begin{table}
	\caption{Fine-tuning schedules and exact match and F1-scores for extractive QA.}
	\label{tab:freq}
	\begin{tabular}{ccc}
		\toprule
		Schedule No.& Exact Match Score& F1-Score\\
		\midrule
    		1 & 32.57 & 42.26 \\
    		2 & 38.25 & 41.07 \\
    		3 & 33.33 & \textbf{60.39} \\
    		4 & \textbf{39.77} & 42.91 \\
		\bottomrule
	\end{tabular}
\end{table}

\begin{table*}
	\caption{Fine-tuning schedules with single and multiple text transformations and model accuracy scores for Yes/No QA.}
	\label{tab:freq}
	\begin{tabular}{ccl}
	\toprule
	Schedule No.& Transformations& Accuracy\\
	\midrule
	Baseline &
	None &
	0.66 \\
	I &
	Single &
	0.66 \\
	II &
	Single &
	\textbf{0.68} \\
	III &
	Single &
	0.66 \\
	IV &
	Single &
	\textbf{0.68} \\
	V &
	Single &
	0.66 \\
	VI &
	Single &
	0.66 \\
	VII &
	Single &
	\textbf{0.69} \\
	VIII &
	Single &
	0.68 \\
	IX &
	Single &
	0.66 \\
	X & Multiple & 0.68 \\
	XI & Multiple & 0.68 \\
	XII & Multiple & \textbf{0.72} \\
	\bottomrule
	\end{tabular}
\end{table*}

Our research goals include attaining a better understanding of the training process of a QA system through variations in the training data. Towards that end, we show how training on data (sub)sets and transformations result in measurable improvement in QA performance compared either to established baselines in the literature or to baselines that we established. Since we do not consider multiple model architectures, we proceed with the caveat that our results would likely not match optimal performance values on any given QA task. Nevertheless, in this section, we illustrate the impact of each of our methods.

To the best of our knowledge, we present the first set of results for extractive QA using PubMedQA \cite{jdlcl:19} in Table 1, thereby establishing a baseline for this task (schedule 2). For exact match, in most cases, our models achieved better performance, as we trained in phases on data that matched the distribution of our target test set. We explain the variations in model performance across the fine-tuning stages by the different characteristics of the QA pairs in the training schedules. Additionally, we evaluate the model trained under schedule 3 on a small hold-out set of approximately 20 parameterized QA pairs not used in training and not included in the target test set, achieving an exact match score=42.85 and $F1$-score=70.13.

We compare our extractive QA results with related work \cite{mlxw:21} on open-domain QA using BioBERT trained on COVID-QA \cite{mrjp:20}, a dataset of 2,019 QA pairs, annotated by biomedical domain experts on scientific articles related to COVID-19. While this type of data would be very useful to have, we assume that expert-annotated data for a new disease may not be readily available at disease onset. Nevertheless, our best exact match score of 39.77, under schedule 4, is exactly in line with the best score of 39.16 using COVID-QA \cite{mlxw:21}. This points to the effectiveness of training an extractive QA model using BM25-based weak supervision along with COVID-specific curated data.

To understand our $F1$-scores better, we examine the characteristics of the curated data. The majority of QA pairs used to fine-tune our models are derived from papers and are not representative of natural questions. This could explain the modest improvement from automatically-curated (artificial) PubMed COVID QA pairs under schedule 4. Only models fine-tuned under schedules 3 and 4 are trained on more than 500 natural, parameterized COVID QA pairs. While the introduction of these pairs (schedule 3) has a negative impact on the exact match score, it results in a significant improvement of the $F1$-score. The resulting model appears to benefit from regularities in the parameterized QA pairs (e.g., multiple answers to each question, structural similarity across questions). 

Further work is needed to tap the potential of natural questions and artificial COVID QA pairs. The path to system deployment includes increasing the number and diversity of parameterized COVID QA pairs to handle variation in clinical questions and training our models on large natural question datasets, including Google's Natural Questions \cite{kprcpaepdltjkcdulp:19}. We also plan to train on additional artificial COVID QA pairs, incrementally downloading new PubMed papers as they become available.

For yes/no QA, by applying single text transformations in schedules II, IV, and VII, in Table 2, we see an improvement in accuracy of 3\% to 4.5\%, compared to the baseline. However, the improvement is marked -- 9\% over the baseline (no lemmatization or subword tokenization/content expansion using neo-classical combining forms) -- when we concatenate lemmas, replace neo-classical forms, and replace form meanings in schedule XII. Our results suggest that single text transformations are complementary, and there is a benefit to combining them. 

For comparison, we look at performance on text data augmentation involving text editing \cite{wz:19}. \citeauthor{wz:19} (\citeyear{wz:19}) used different models and datasets, so it is challenging to do a fair comparison. However, for perspective, their data augmentation techniques resulted in an average percentage increase in accuracy on text classification tasks using CNN and RNN models of 3.9\% on small datasets and 0.9\% on full datasets. 

Prior to deploying yes/no models in our QA system, we need to curate COVID-specific data. Our parameterized COVID QA data curation process can be adapted readily for yes/no QA. \citeauthor{jdlcl:19}  (\citeyear{jdlcl:19}) provided a recipe for automatically creating ``noisy" yes/no QA pairs. Then, we would transform the text of the curated data as described above and replicate fine-tuning schedules 1-4 for yes/no QA.

\section{Conclusions}
In this paper, we show how varying training data improves deep neural network model performance for two question answering tasks. We train an extractive QA model using BM25-based weak supervision with labels generated automatically from repurposed structured article abstracts and show BM25 labelling to be a viable alternative to human ground-truth labelling. We curate COVID-specific natural and artificial questions using structured data and information retrieval techniques that did not require annotation by biomedical domain experts. We show how to stage extractive QA model fine-tuning to achieve domain specialization. For COVID-specific questions, we show how to stage model fine-tuning from easier, natural questions to harder, artificial questions. In both cases, we measure and analyze differences in model performance after each stage. For yes/no QA, by augmenting our training data with linguistic features to account for variations in word morphology and meaning, we show that how we represent textual content matters for model performance.

\begin{acks}
This research is based upon work supported in part by the Office of the Director of National Intelligence (ODNI), Intelligence Advanced Research Projects Activity (IARPA), via Contract \#: 2021-20120800001. The views and conclusions contained herein are those of the authors and should not be interpreted as necessarily representing the official policies, either expressed or implied, of ODNI, IARPA, or the U.S. Government. The U.S. Government is
authorized to reproduce and distribute reprints for governmental purposes notwithstanding any copyright annotation therein.
\end{acks}

\bibliographystyle{ACM-Reference-Format}
\bibliography{Arxiv_PeratonLabs_Submission}

\end{document}